%% file: acl_latex.tex
\pdfoutput=1
\documentclass[11pt]{article}

\usepackage[final]{acl}
\usepackage{graphicx}
\usepackage{times}
\usepackage{comment}
\usepackage{pifont}
\usepackage{makecell}
\usepackage{hyperref}
\usepackage{amssymb}
\usepackage{multirow}
\usepackage{latexsym}
\usepackage{tabularx}
\usepackage{booktabs}
\usepackage{amsmath}
\usepackage{times}
\usepackage{latexsym}

\usepackage[T1]{fontenc}
\usepackage[utf8]{inputenc}

\usepackage{microtype}

\title{MetaSumPerceiver: Multimodal Multi-Document Evidence Summarization for Fact-Checking}

\author{
  Ting-Chih Chen \\
  Virginia Tech \\
  \texttt{tingchih@vt.edu} \\\And
  Chia-Wei Tang \\
  Virginia Tech \\
  \texttt{cwtang@vt.edu} \\\And
  Chris Thomas \\
  Virginia Tech \\
  \texttt{chris@cs.vt.edu} \\
  }

\begin{document}
\maketitle
\begin{abstract}

Fact-checking real-world claims often requires reviewing multiple multimodal documents to assess a claim's truthfulness, which is a highly laborious and time-consuming task. In this paper, we present a summarization model designed to generate claim-specific summaries useful for fact-checking from multimodal, multi-document datasets. The model takes inputs in the form of documents, images, and a claim, with the objective of assisting in fact-checking tasks. We introduce a dynamic perceiver-based model that can handle inputs from multiple modalities of arbitrary lengths. To train our model, we leverage a novel reinforcement learning-based entailment objective to generate summaries that provide evidence distinguishing between different truthfulness labels. To assess the efficacy of our approach, we conduct experiments on both an existing benchmark and a new dataset of multi-document claims that we contribute. Our approach outperforms the SOTA approach by 4.6\% in the claim verification task on the MOCHEG dataset and demonstrates strong performance on our new Multi-News-Fact-Checking dataset. \href{https://github.com/tingchihc/metasumperceiver}{Code}

\end{abstract}

\section{Introduction}
\input{introduction}

\section{Related Work}
\input{related_work}

\section{Multi-News-Fact-Checking Dataset} \label{se:MNFCS}
\input{dataset}

\section{Approach}

\input{approach}

\section{Experiments}
\input{experiments}

\section{Conclusion}

\input{conclusion}

\section{Acknowledgements}
We acknowledge Advanced Research Computing at Virginia Tech for providing computational resources and technical support that have contributed to the results reported within this paper. We also thank all reviewers for their comments which helped improve the paper.

\clearpage

\section{Limitations}
\input{limitation}

\bibliography{anthology,custom}
\bibliographystyle{acl_natbib}

\clearpage
\section{Appendix}
\input{appendix}

\end{document}

%% file: introduction.tex
\begin{figure}
        \hspace*{-0.5cm}\includegraphics[width=0.6\textwidth,height=0.4\textwidth,keepaspectratio]{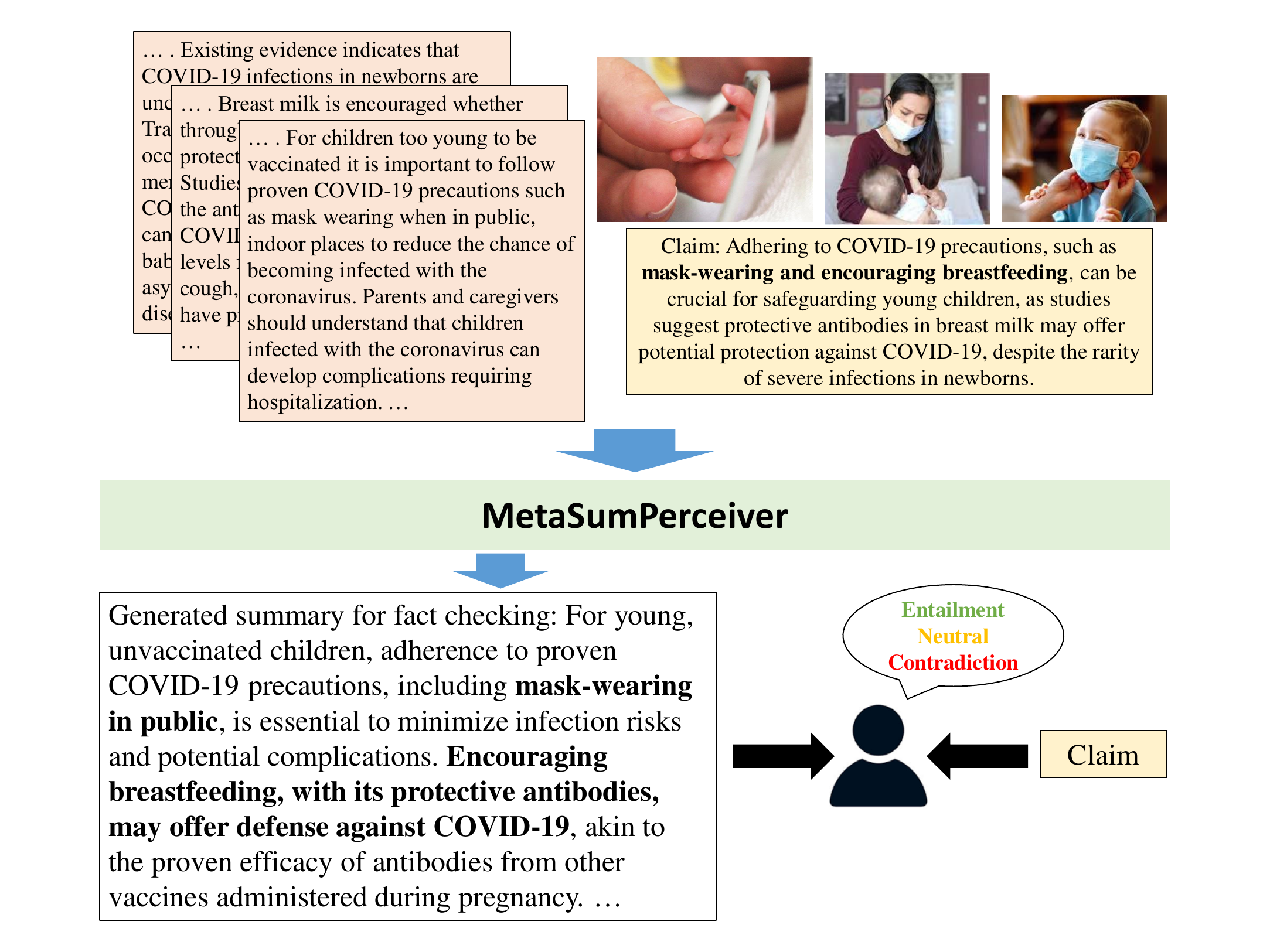}
  % \vspace{-0.25em}
  \caption{Overview of MetaSumPerceiver (MSP): Using inputs such as documents, images, and claims, MSP generates summaries to facilitate fact-checking. In this example, the summary provides evidence and establishes that the claim in question is entailed by the evidence.}
  \label{fig:metasumperceiver_overview}
  \vspace{-1em}
\end{figure}

\begin{table*}[ht]
\centering
\caption{Comparing Multi-News-Fact-Checking with other fact-checking datasets. Topic(s) is the inferred value.}
\resizebox{\linewidth}{!}{
\begin{tabular}{l|cccccccccc}
\hline
\textbf{Datasets}& \textbf{\#Samples}& \textbf{Source}& \textbf{Topic(s)} & \textbf{Language}& \textbf{Multi-modal}& \textbf{Multi-doc}& \textbf{Verification}& \textbf{Explanations}& \textbf{Text/image retrieval} \\ \hline

~\citet{zlatkova2019factchecking}& 1,233& Snopes, Reuters& < 1,500& English& \ding{52}& \ding{55}& \ding{52}& \ding{55}& \ding{55}\\
~\citet{cheema2022mmclaims}& 3,400& X& < 3,400 & English& \ding{52}& \ding{52}& \ding{52}& \ding{55}& \ding{52}\\
~\citet{nielsen2022mumin}& 12,914& X& 26,048 & Multi& \ding{52}& \ding{52}& \ding{52}& \ding{55}& \ding{55}\\
~\citet{Yao_2023}& 15,601& Politifact, Snopes& < 15,631 & English& \ding{52}& \ding{55}& \ding{52}& \ding{52}& \ding{52}\\
~\citet{nakov2021overview}& 18,014& X& < 1,312& Multi& \ding{55}& \ding{52}& \ding{52}& \ding{55}& \ding{55}\\
Ours  & 111,905& Multi-News& < 1,500& English& \ding{52}& \ding{52}& \ding{52}& \ding{52}& \ding{52}\\
\hline
\end{tabular}
}
\label{label:compare}
\vspace{-0.5em}
\end{table*}

Fact-checking claims on social media platforms poses a significant challenge due to the large volume of new claims constantly being posted without sufficient methods for verification~\cite{Aimeur2023-cb}. Research indicates that manually verifying all aspects of a 200-word claim can require up to four hours of dedicated effort~\cite{vladika2023scientific}. Further, despite the exceptional capabilities of large language models (LLMs) in natural language processing tasks, they still generate content with factual errors. Given that LLMs can produce convincing statements and thus influence beliefs, the potential for hallucinations poses a serious risk of misleading users when deployed for fact-checking~\cite{Jakesch_2023, Jakesch_2023_1, kreps_mccain_brundage_2022}. This concern highlights the possibility of language models becoming new sources of misinformation and disinformation. The proliferation of misinformation and fake news adds to this predicament, making it increasingly difficult to distinguish between reliable and deceptive content~\cite{goldstein2023generative, spitale2023ai}. Thus, there is an urgent need for tools capable of succinctly summarizing relevant evidence for fact-checkers, i.e., systems that provide a brief yet comprehensive overview of the relevant evidence to facilitate accurate and reliable assessments. Existing research relying on summarization for fact-checking is ineffective because these methods fail to extract evidence from the resources~\cite{Das_2023, doi:10.1177/14614448211012377, berlinski_doyle_guess_levy_lyons_montgomery_nyhan_reifler_2023}.

A potential solution to this problem is provided by multimodal summarization~\cite{khullar2020mast, liu2023long}, which can generate summaries from sources including text, images, videos, and audio. This is a challenging task because each modality might contribute complementary information, e.g., a bar chart image accompanying relevant facts mentioned in the text. Present methods usually produce summaries given a limited number of inputs~\cite{puduppully_data--text_2021, wang-etal-2022-robust}. The research challenge lies in processing arbitrary inputs from diverse modalities and discerning the explicit relationships between them. However, unlike the standard summarization task, which seeks to summarize the salient content of an article, our objective is to effectively distill claim-specific evidence useful for fact-checking across various modalities.

To train our system to generate summaries useful for human fact-checking, we assess the utility of our summaries at performing entailment~\cite{inproceedings}, a closely aligned task to fact-checking. Our work is orthogonal to prior work in entailment, in that rather than learning to predict the entailment label for the premise-hypothesis pair, we seek to generate the premise for a specific claim from a pool of multimodal data. To address the limitations of applying existing summarization methods for fact-checking, we propose the~\textbf{MetaSumPerceiver} (MSP) model in Figure~\ref{fig:metasumperceiver_overview}, where the input consists of a claim, a set of documents and images, and the objective is to generate a summary that expedites the fact-checking process for humans. We initially train the perceiver model with a summarization model. Subsequently, to produce the summary for fact-checking, we employ a proxy reward mechanism to update the summarizer to ensure the generation of an accurate and relevant summary with necessary evidence.

To support research on the task of multi-document fact-checking, we contribute a benchmark (Multi-News-Fact-Checking) of claims and entailment labels whose evidence is drawn from multiple documents. We evaluate our method on the MOCHEG benchmark~\cite{Yao_2023} and our new dataset and demonstrate substantial improvements compared to existing baselines. The major contributions of this paper are as follows:

\begin{itemize}
  \item We present an innovative approach for multimodal multi-document summarization specifically designed for fact-checking applications.

  \item We release the Multi-News-Fact-Checking dataset, to support the multimodal multi-document fact-checking summarization task.
  
  \item We perform detailed experiments and ablations of our model and loss functions which clearly demonstrate the superiority of our approach over existing methods.
  
\end{itemize}

%% file: related_work.tex
\subsection{Perceiver}

The Perceiver architecture~\cite{jaegle2021perceiver} enables scaling transformers to input sequences of arbitrary lengths, by reducing the memory footprint in standard self-attention. Follow-up works, such as Perceiver IO~\cite{jaegle2022perceiver}, adapt the original model by presenting a versatile architecture adept at processing data from various settings while ensuring linear scalability with input and output dimensions. The model has demonstrated strong performance on many downstream tasks, including optical flow estimation~\cite{Butler:ECCV:2012} and the GLUE language benchmark~\cite{wang2019glue}. Our method relies on~\citet{jaegle2022perceiver} to process a variable number of arbitrarily long text documents and images. We use the model in sequence with a summarization model to generate a multimodal summary.

\subsection{Multimodal Fact-checking Datasets}

In the current task of fact-checking using multiple datasets~\cite{zlatkova2019factchecking, nakov2021overview, cheema2022mmclaims, nielsen2022mumin, Yao_2023}, the main sources of data are X and Snopes, with a focus on COVID-19, elections, and the Russo-Ukrainian war. This has led to a couple of issues. First, X has already blocked their API, making it difficult for people to access these datasets. Second, we are seeking a dataset that covers a variety of topics rather than being limited to specific ones. Additionally, we prefer a dataset in the English language. To address these issues, we propose a new multimodal fact-checking dataset based on Multi-News~\cite{alex2019multinews}, which includes information from multiple documents and related images, covering a broader array of topics such as news, policy, weather, sports, etc. Table~\ref{label:compare} provides a comparison of the Multi-News-Fact-Checking dataset with the aforementioned datasets.

\subsection{Learning From Feedback}

Recent advancements in LLMs have revolutionized the AI landscape ~\cite{touvron2023llama1, touvron2023llama, driess2023palme, openai2023gpt4}. However, because they are mostly trained on data scraped from the web LLMs sometimes produce undesired outcomes, including generating biased or harmful content~\cite{10.1145/3442188.3445922}. Recognizing the importance of aligning LLMs with human values, has led to efforts in supervised fine-tuning (SFT) with ethical guidelines~\cite{alpaca}. While these efforts demonstrate the potential of integrating human feedback into training using reinforcement learning for user-tailored tasks~\cite{ouyang2022training, bai2022training}, training LLMs to reflect human values is quite challenging. In our work, we adopt the idea of training language models with feedback. However, rather than relying on a human fact-checker, we utilize a surrogate reward model (an entailment model) to stand in the place of a human fact checker, in order to fine-tune the summarizer to generate summaries that give evidence for fact-checking specific claims through Proximal Policy Optimization (PPO)~\cite{schulman2017proximal, zheng2023secrets}.

\begin{comment}
\begin{table}[h]
\centering
\caption{Accuracy for verifying the entailment, neutral, contradiction claims with Llama-2-70b in Multi-News-Fact-Checking dataset.}
\begin{tabular}{l|c}
\hline
\textbf{Claims}      & \textbf{Accuracy(\%)} \\ \hline
Entailment claims    & 78.3          \\
Nentral claims       & 64.2          \\
Contradiction claims & 74.1          \\ \hline
\end{tabular}
\label{label:prompt_accuracy_llama2}
\end{table}
\end{comment}

\begin{table}[h]
\centering
\vspace{-0.25em}
\caption{Analysis of claims in the Multi-News-Fact-Checking dataset. Top: Entailment accuracy using Llama 2. Bottom: Classification results indicating claim checkworthiness.}
\vspace{-0.25em}
\resizebox{\linewidth}{!}{
\begin{tabular}{l|c}
\hline
\textbf{Entailment label consistency} & \textbf{Accuracy (\%)}\\\hline
Entailment claims    & 78.3 \\
Nentral claims       & 64.2 \\
Contradiction claims & 74.1 \\\hline
\textbf{Checkworthiness results} & \textbf{Percentage (\%)}\\\hline
Unimportant factual sentence (UFS) & 17.67 \\
Checkworthy factual sentence (CFS) & 68.6 \\
Non-factual sentence (NFS) & 13.71 \\\hline
\end{tabular}
}
\label{label:test111}
\vspace{-0.75em}
\end{table}

%% file: dataset.tex
To train our system, we need a dataset of claims whose facts are drawn from multiple documents along with the entailment label of each claim. We build our dataset on top of the Multi-News summarization dataset~\cite{alex2019multinews}, which contains sets of multiple text documents along with human-written summaries of each set. Because the Multi-News dataset lacks claims specifically tailored for fact-checking tasks, we prompt Llama 2~\cite{touvron2023llama} to generate labeled claims from each set of documents. Within each group of Multi-News documents, we leverage the human-written multi-document summary to generate 30 claims (ten of each entailment type), resulting in a dataset of 1,291,168 labeled claims. The specific prompts contain sections containing a task description, example, and instructions, which are fully detailed in the appendix~\ref{app:prompts}. Our dataset contains 111,905 images we obtained by retrieving images from the original articles.

To assess the quality and effectiveness of claim generation, we conducted an evaluation using a scale from 1 to 5, where 1 indicates low quality and 5 indicates high quality. We tested 60 claims and obtained an average score of 3.61 for claim generation quality and effectiveness. In comparison, the claims generated from PRIMERA summaries scored an average of 3.21. To ensure impartiality, the annotators were blinded and asked to rate the claims alongside their respective articles and summaries. 

Additionally, in Table~\ref{label:test111}, we validated claim labels (entailment, neutral, and contradiction) using Llama 2. Specifically, we treated the ground truth label (i.e., the label used to prompt Llama 2 to generate a claim) as the ground truth and prompted Llama 2 in a zero-shot manner to predict the entailment labels. The average accuracy for claim verification was 72.2\%, indicating that the generated claims were largely consistently predicted as their intended labels.

\begin{comment}
\begin{table}[h]
\centering
\caption{Classification results indicating whether the claims are deemed check-worthy or not.}
\begin{tabular}{l|c}
\hline
\textbf{Classes}      & \textbf{Accuracy(\%)} \\ \hline
UFS & 17.67          \\
CFS & 68.6          \\
NFS & 13.71          \\ \hline
\end{tabular}
\label{label:claimbuster}
\end{table}
\end{comment}

To assess the checkworthiness of our generated claims (i.e., to ensure that they are factual claims worth fact-checking), we use a pre-trained model trained on the ClaimBuster dataset~\cite{arslan2020benchmark}, as illustrated in Table~\ref{label:test111}. The model assigns claims into three classes: UFS (unimportant factual claims that are not considered check-worthy), CFS (claims containing factual information of public interest in terms of their veracity), and NFS (sentences that do not contain any factual claims). 
% UFS denotes factual claims that are not considered check-worthy. CFS indicates claims containing factual information of public interest in terms of their veracity. NFS signifies sentences that do not contain any factual claims. 
The result shows that 70\% of the prompted claims are check-worthy claims. This outcome substantiates that our prompts are well-designed for this task, and that Llama 2 accurately comprehend the task's intended meaning without misunderstanding.

\begin{figure*}
  \centering
  \includegraphics[width=\textwidth, height=0.4\textwidth, keepaspectratio]{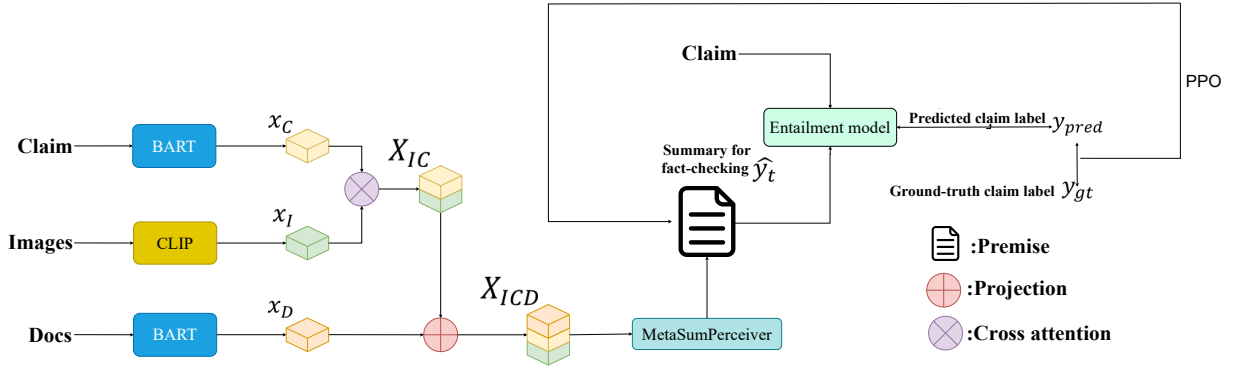}
  \caption{Overview of MetaSumPerceiver (MSP): This figure illustrates the process of generating a summary for fact-checking using MSP, integrating a fixed entailment model for accurate truthfulness labeling. Furthermore, it highlights how PPO is employed to continually refine the summary during the fact-checking process.}
  \label{fig:metaover}
\end{figure*}

%% file: approach.tex
In this section, we explain the details of our approach, MSP as illustrated in Figure~\ref{fig:metaover}. We also describe the preprocessing steps for both text and image data, the components of our model, and the reinforcement learning methodology we applied to train MSP. Our approach is capable of summarizing multiple multimodal documents consisting of arbitrarily long texts and images. Specifically, we use $x_C$, $x_D$, and $x_I$ to represent embeddings for claims, documents, and images, respectively.

\subsection{Preprocessing}

Due to the sequence length limitation, we utilize a combination of Perceiver with BART~\cite{lewis2019bart} and CLIP~\cite{radford2021learning} to extract embeddings. To circumvent exceeding the sequence length, we break down the data into chunks of 1024 tokens. Subsequently, we employ Perceiver to merge these chunks for both textual and visual embeddings. For the textual data, we use BART to obtain text embeddings following~\cite{devlin-etal-2019-bert}. As a result, each input text is transformed into a set of token embeddings $x_C \in \mathbb{R}^{n \times D}$ and $x_D \in \mathbb{R}^{m \times D}$, where $n$ and $m$ are the number of tokens and $D$ is the dimension of embedding. Then, we use CLIP (ViT-G-14) to extract visual features for the images. Finally, each input image undergoes a transformation, resulting in a set of visual embeddings. $x_I \in \mathbb{R}^{k \times D}$, where $k$ is the number of tokens and $D$ is the dimension of the embedding.
 
\subsection{Model Training Strategy}

Our goal is to generate a textual summary of a set of multimodal documents that enables a fact-checker to determine the veracity of a claim. In order to select relevant visual content from the images, we begin by performing a cross-attention between the images and the claim:

\begin{equation}
  \begin{array}{l}
    X_{IC}= ATTN(Q_{x_C}, K_{x_I}, V_{x_I}) \;,
  \end{array}
\end{equation}

where the query $Q_{x_C}$ is the claim's sequence of embeddings and $K_{x_I}$ and $V_{x_I}$ are the embedding sequences of visual tokens from the images. We project $X_{IC}$ into the document embedding $X_D$, which serves as the input for MSP. The output from the cross-attention block, $X_{IC}$, is initially projected by a linear projection layer with the weight $\theta$. It is then concatenated with $x_D$, as depicted in the subsequent equation:

\begin{equation}
  \begin{array}{l}
  X_{ICD} = 
  \begin{bmatrix}
    proj(X_{IC}, \theta)^\intercal , X_D ^\intercal 
  \end{bmatrix}^\intercal,
  \end{array}
\end{equation}

where $X_{ICD}$ will be the input to MSP. Prior to training our full model, we pre-train our attention block and summarization model using the Multi-News dataset's human written summaries using the cross-entropy loss function:

\begin{equation}
  \begin{array}{l}
    \mathcal{L}_{\text{sum}} = -\sum_{t=1}^{T} \sum_{i=1}^{N} y_{t_i} \log(\hat{y}_{t_i}),
  \end{array}
\end{equation}

where $T$ represents the sequence length, $N$ is the vocabulary size, and $y_{t_i}$ and $\hat{y}_{t_i}$ denote the ground truth and predicted probabilities of token $i$ at time step $t$, respectively. In the remaining text, we omit the summation over the vocabulary for conciseness.

\begin{figure*}
  \centering
  \includegraphics[width=\textwidth, height=0.4\textwidth, keepaspectratio]{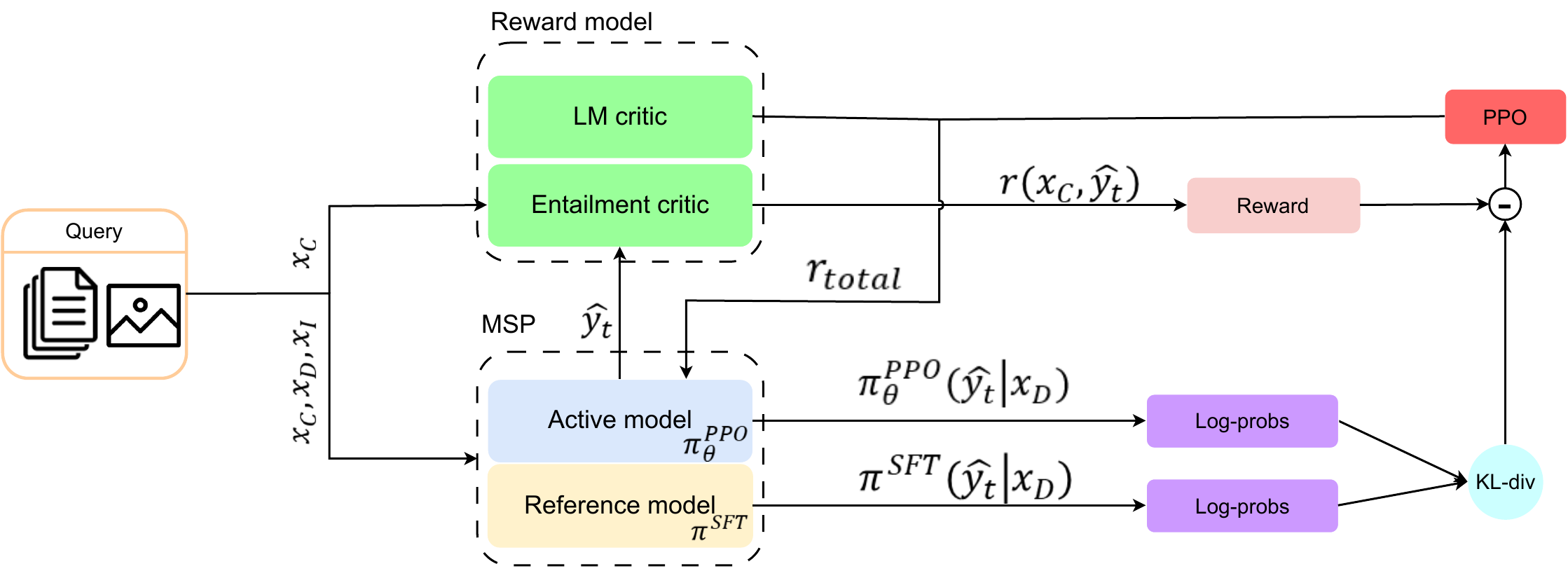}
  \caption{The Proximal Policy Optimization (PPO) process starts with the summarizer generating a response based on the input query. The reward model then assesses this query-response pair, producing a single reward score. Simultaneously, the process calculates the KL-divergence by comparing the likelihood of token sequences in the response with both the currently fine-tuned active model and a pre-trained reference model. The KL-divergence acts as a measure of reward, ensuring that responses from the active model align with those from the reference model. Additionally, we input the summary into Mistral LM to evaluate whether the summary is concise or not. In conclusion, PPO updates the parameters of the active model based on the reward model's output, Mistral LM, and the value of the KL-divergence.}
  \label{fig:rl_ppo}
\end{figure*}

\subsection{Fine-tuning The Summarizer}
To enhance the summarizer's ability to produce summaries that provide the evidence needed for fact-checking claims, we adopt the concept of training a language model using feedback with reinforcement learning. After pretraining the perceiver and summarization models, we employ reinforcement learning with an entailment model serving as a surrogate for a human fact-checker as feedback. We first exclusively apply reinforcement learning to the perceiver. Subsequently, we unfreeze the summarizer and continue training end-to-end with both the perceiver and summarizer. 
We illustrate our fine-tuning process in Figure~\ref{fig:rl_ppo}.

\subsubsection{Reward Model For Fact-Checking}

Contrary to the approach in reinforcement learning from human feedback, which necessitates a human arbitrator to score the model's outputs, in this study, we train a reward model to act like a human fact-checker to guide the summarizer in producing summaries for fact-checking instead. We utilized a comprehensive dataset consisted with MultiNLI~\cite{N18-1101}, Fever-NLI~\cite{Thorne18Fever}, and Adversarial-NLI (ANLI)~\cite{nie2020adversarial}, encompassing a total of 763,193 premise-claim pairs. Leveraging this dataset, we fine-tuned DeBERTAV3~\cite{he2023debertav3} for the task of entailment classification using cross-entropy loss. Serving as an entailment classifier, this model achieves accuracy rates of 90.3\%, 77.7\%, and 57.9\% in the MultiNLI, Fever-NLI, and ANLI evaluation datasets, respectively.

\subsubsection{Proximal Policy Optimization}

We define the score from the reward model as the probability of the ground-truth label given both the claim (as the hypothesis) and the generated summary for fact-checking (as the premise). The formulation for the score from the reward model can be formulated as:

\begin{equation}
  \begin{array}{l}
    r(x_C, \hat{y}_t) = P(y_{gt}|{x_C, \hat{y}_t}) - \\
    \\
    0.5 * \Sigma_{y_{gt} \neq y_{pred}} P(y_{pred}|{x_C, \hat{y}_t}),
  \end{array}
\end{equation}

where $x_C$, $\hat{y}_t$, $y_{gt}$ and $y_{pred}$ denote the claim, the generated summary, the groud-truth label of the claim, and the predicted label of the claim, respectively. The value of $P(y_{\{gt,pred\}}|x_C,\hat{y}_t)$ is derived from the trained entailment classifier. The primary objective behind this reward function is to maximize the likelihood that the generated summary for fact-checking contains the facts necessary for the model to predict the claim's ground truth label.

In this paper, we employ PPO as our policy gradient method for reinforcement learning. PPO adds an additional term to the reward function, which imposes a penalty determined by the Kullback-Leibler (KL) divergence between the trained RL policy summarizer, $\pi^{PPO}_{\phi}$, and the initial supervised summarizer $\pi^{SFT}$.

\begin{table*}[!th]
\centering
\caption{ Performance of claim verification in MOCHEG with our method. We separately calculate the precision and recall in supported, refuted, and NEI claim labels. We compare our method with published baselines in Table~\ref{label:claim_verification_fscore_MOCHEG}. The labels "Supported," "NEI," and "Refuted" in fact-checking classification are analogous to truthfulness labels. "Supported" aligns with the entailment label, indicating that the hypothesis is similar to the premise. "NEI" corresponds to "not enough information" and is comparable to the neutral label, indicating that the hypothesis includes both information entailed by the premise and information that lacks clarity or confirmation. "Refuted" shares the same classification as contradiction, indicating that the hypothesis is not entailed with the premise.}
\resizebox{\textwidth}{!}{
\begin{tabular}{l|ccccccc}\hline
\textbf{Setting}      & \textbf{Accuracy (\%)} & \textbf{\begin{tabular}[c]{@{}c@{}}Precision (\%)\\ Supported\end{tabular}} & \textbf{\begin{tabular}[c]{@{}c@{}}Precision (\%)\\ Refuted\end{tabular}} & \textbf{\begin{tabular}[c]{@{}c@{}}Precision (\%)\\ NEI\end{tabular}} & \textbf{\begin{tabular}[c]{@{}c@{}}Recall (\%)\\ Supported\end{tabular}} & \textbf{\begin{tabular}[c]{@{}c@{}}Recall (\%)\\ Refuted\end{tabular}} & \textbf{\begin{tabular}[c]{@{}c@{}}Recall (\%)\\ NEI\end{tabular}} \\
\hline
MSP (Entail.~critic) w/ Text Evidence $\rightarrow$ DeBERTAV3  & 43.7 & 79.2 & 66.9 & \textbf{33.9} & 40.5 & 30.6 & 25.8\\
MSP (Entail.~critic) w/ Text + Img Evidence $\rightarrow$ DeBERTAV3 & 50.8 & 83.4 & \textbf{69.3} & 27.3 & 42.9 & 34.2 & 30.9\\
MSP (Entail.~critic) w/ Text Evidence $\rightarrow$ Llama 2  & 46.7 & 80.4 & 68.1 & 31.5 & 37.2 & 35.4 & 31.5\\
MSP (Entail.~critic) w/ Text + Img Evidence $\rightarrow$ Llama 2 & \textbf{53.7} & \textbf{87.3} & 60.3 & 32.4 & \textbf{48.3} & \textbf{36.9} & \textbf{34.8} \\ \hline

MSP (Entail., LM critics) w/ Text Evidence $\rightarrow$ DeBERTAV3  & 40.2 & 77.3 & 63.4 & \textbf{45.9}& 38.2 & 35.7 & 28.4 \\
MSP (Entail., LM critics) w/ Text + Img Evidence $\rightarrow$ DeBERTAV3 & 47.8 & 78.1 & \textbf{67.5} & 38.1 & 39.5 & \textbf{37.5} & 34.1\\
MSP (Entail., LM critics) w/ Text Evidence $\rightarrow$ Llama 2  & 49.3 & 81.5 & 65.2 & 37.4 & 39.7 & 31.5 & 35.7\\
MSP (Entail., LM critics) w/ Text + Img Evidence $\rightarrow$ Llama 2 & \textbf{55.6} & \textbf{88.2} & 57.5 & 39.6 & \textbf{51.2} & 32.4 & \textbf{37.2}
\\\hline
\end{tabular}}
\label{label:MOCHEG}
\end{table*}

Moreover, we incorporate an extra reward $r_{quality}$, LM critic, to evaluate the quality of the summary, specifically focusing on clarity and conciseness. We utilize Mistral~\cite{jiang2023mistral} along with a detailed quality testing prompt provided in the appendix to assess this aspect. The assigned reward ranges from 0 to 1. We integrate $r_{quality}$ into our model update in conjunction with the existing $r_{total}$. The cumulative reward is described as follows:

\begin{equation}
  \begin{array}{l}
    r_{total} = (r_{quality} + r(x_C, \hat{y}_t) - \\
    \\
    \eta KL(\pi^{PPO}_{\phi}(\hat{y}_t|x_D), \pi^{SFT}(\hat{y}_t|x_D)))/2,
  \end{array}
\end{equation}

where $\eta$ represents the KL reward coefficient, which determines the magnitude of the KL penalty, we set it to 0.2 for our model. This coefficient functions as an entropy boost, enhancing exploration throughout the policy domain and urging the model to engage in a diverse set of actions rather than the one currently considered the best. In addition, it inhibits the policy from rapidly committing to a singular strategy, and this encourages outputs from the RL fine-tuned model to not deviate too far from the original model. 

MSP is optimized through PPO based on the policy gradient methods that optimize the policy of the model using gradient ascent. The update rule for the policy gradient is given as: 

\begin{equation}
  \begin{array}{l}
    \theta \longleftarrow \theta + \alpha \nabla_{\theta} J(\theta),
  \end{array}
\end{equation}

where $\alpha$ and $J_{\theta}$ denote the learning rate and the expected return under policy $\pi_{\theta}$ from the model, respectively.

%% file: experiments.tex
\subsection{Claim Verification}

The goal of our method is to generate a summary from multiple documents and modalities that is useful for fact-checking a claim. In order to assess how useful our method is at this task, we compare the performance of our method on MOCHEG, which presents a benchmark and method for multi-document multimodal fact-checking. Specifically, we employed two~\textit{fixed} entailment models, namely DeBERTAV3~\cite{he2023debertav3} and Llama 2~\cite{touvron2023llama}, as our surrogate ``human'' fact checkers to predict the entailment label of a claim given our generated summary. Importantly, we do not fine-tune these models with our generated summaries to avoid biasing the models toward the linguistic or stylistic patterns of the summaries. This ensures that they do not learn spurious features in the downstream task.

As depicted in Tables~\ref{label:MOCHEG} and ~\ref{label:claim_verification_fscore_MOCHEG}, our method exhibits superior performance, achieving a SOTA 48.6 F-score in the MOCHEG dataset. Furthermore, according to Table~\ref{label:MOCHEG}, our method demonstrates strong precision performance for the ''Supported'' label. We conduct separate tests with two distinct critics. It is noted that the most optimal performance was achieved when deploying both the entailment critic and the LM critic. This outcome indicates that our model is proficient in verifying claim labels through clear and concise summaries.

Table~\ref{label:MOCHEG} reveals that the best results are achieved when inputs incorporate both textual and image evidence. Perhaps unsurprisingly given its size, the zero-shot Llama 2 entailment surrogate model surpasses DeBERTAV3 in performance. Nevertheless, a notable issue persists, where the surrogate entailment models struggle to accurately deal with NEI claim labels.

Table~\ref{label:claim_verification_fscore_MOCHEG} highlights the superiority of our model compared to MOCHEG. In the case of MOCHEG, truthfulness labels are predicted by averaging a stance representation derived from both textual and image evidence. Furthermore, MOCHEG's classifier relies on fixed thresholds, which may not be optimal for every situation. In contrast, our approach involves generating summaries for fact-checking via reinforcement learning with fixed entailment models and LM critic. Although a difference remains in the result of human vs system prediction performance, our model surpasses the prior state-of-the-art system by 4.6\% F-score.

\begin{table}[!th]\footnotesize
\centering
\caption{Performance of claim verification in MOCHEG. DeBERTaV3 and Llama 2 represent the fixed entailment models. "Gold Evidence" denotes ground truth text and image evidence, while "System Evidence" refers to automatically retrieved text and image evidence. "Human" indicates human evaluation.}
\resizebox{\linewidth}{!}{
\begin{tabular}{l|c}
\hline
\textbf{Setting}                       & \textbf{F-score (\%)} \\ \hline
MSP(Entail.~critic) w/ Text Evidence $\rightarrow$ DeBERTAV3           & 42.7                   \\
MSP(Entail.~critic) w/ Text + Img Evidence $\rightarrow$ DeBERTAV3 & 45.1                   \\
MSP(Entail.~critic) w/ Text Evidence $\rightarrow$ Llama 2           & 43.9                 \\
MSP(Entail.~critic) w/ Text + Img Evidence $\rightarrow$ Llama 2 & \textbf{48.2}                   \\ \hline

MSP(Entail., LM critics) w/ Text Evidence $\rightarrow$ DeBERTAV3           & 44.1                  \\
MSP(Entail., LM critics) w/ Text + Img Evidence $\rightarrow$ DeBERTAV3 & 46.1                   \\
MSP(Entail., LM critics) w/ Text Evidence $\rightarrow$ Llama 2           & 44.6                \\
MSP(Entail., LM critics) w/ Text + Img Evidence $\rightarrow$ Llama 2 & \textbf{48.6}                   \\ \hline

MOCHEG w/ Text Evidence                 & 42.7                \\
MOCHEG w/ Image Evidence                & 40.9                \\
MOCHEG w/ Text and Image Evidence       & \textbf{44.0}                \\ \hline

Human w/o Evidence                      & 20.0                \\
Human w/ System Evidence                & 62.0                \\
Human w/ Gold Evidence                  & \textbf{70.0}                \\ \hline
\end{tabular}
}
\label{label:claim_verification_fscore_MOCHEG}
\end{table}

\subsection{Ablations}

Additionally, we conducted ablation experiments for claim verification on our Multi-News-Fact-Checking dataset. A comparative analysis of our method with Llama 2 and other offline summarization models, PRIMERA~\cite{xiao2022primera}, PEGASUS~\cite{zhang2020pegasus} and T5 large~\cite{2020t5}, is presented in Tables ~\ref{label:multi-news_fscore} and ~\ref{label:multi-news}.

Similar to our results in MOCHEG, Tables~\ref{label:multi-news_fscore} and ~\ref{label:multi-news} show that our approach, when employing the Llama 2 surrogate entailment model, achieves the best performance. Furthermore, we achieve balanced accuracy in both precision and recall, underscoring our method's ability to clearly differentiate between truthful and untruthful labels without bias in predictions. The results highlight the inability of other summarization models to generate summaries useful for fact-checking, which causes the surrogate model difficulty in accurately assessing the truthfulness labels. Furthermore, it is evident that the LM critic significantly aids the entailment model in verifying claim labels effectively. The LM critic ensures that the summary is more concise and clear while retaining the essential meaning in the summary. 

In addition, to assess the degree to which our summaries are merely extractive of the source articles, we employed ROUGE to evaluate our summaries alongside the provided articles. Our summaries received a ROUGE score of 0.53, whereas the human-written summaries in the Multi-news dataset scored 0.62. 
Upon comparison, we believe our summaries do not simply rephrase the source articles.

Moreover, to determine the fidelity and informativeness of the generated summaries, we conducted a human evaluation using a scale from 1 to 5, with 1 indicating low fidelity/informativeness and 5 indicating high fidelity/informativeness. We tested 60 generated summaries and obtained an average score of 3.77 for fidelity and informativeness. Comparatively, the PRIMERA summaries scored 3.35. 
To ensure impartiality, we blinded the annotators and asked them to rate the generated summaries alongside their respective articles.

\begin{table}[ht]\footnotesize
\centering
\caption{Performance of claim verification in Multi-News-Fact-Checking dataset. DeBERTAV3 and Llama 2 serve as the fixed entailment models. "Gold Evidence" denotes ground truth text and image evidence, while "System Evidence" refers to automatically retrieved text and image evidence. "Human" indicates human evaluation.} 
\resizebox{\linewidth}{!}{
\begin{tabular}{l|c}
\hline
\textbf{Setting}             & \textbf{F-score (\%)} \\\hline
PEGASUS $\rightarrow$ DeBERTAV3           & 25.4                 \\
PEGASUS $\rightarrow$ Llama 2           & \textbf{30.8}            \\\hline 
T5 large $\rightarrow$ DeBERTAV3          & 28.5            \\
T5 large $\rightarrow$ Llama 2          & \textbf{32.7}            \\\hline 
PRIMERA $\rightarrow$ DeBERTAV3 & 38.2\\
PRIMERA $\rightarrow$ Llama 2   & \textbf{38.3} \\\hline 
MSP(Entail., LM critics) $\rightarrow$ DeBERTAV3               & 40.1            \\
MSP(Entail. critic) $\rightarrow$ Llama 2               & 41.8                 \\
MSP(Entail., LM critics) $\rightarrow$ Llama 2               & \textbf{43.7}        \\\hline
Human w/o Evidence           & 23.0            \\
Human w/ System Evidence     & 65.0               \\
Human w/ Gold Evidence       & \textbf{76.0}               \\ \hline
\end{tabular}}
\label{label:multi-news_fscore}
\end{table}

\begin{table*}[!th]
\centering
\caption{Performance of claim verification in Multi-News-Fact-Checking dataset. We compare our method with Llama 2, and other offline summarization models.}
\resizebox{\textwidth}{!}{
\begin{tabular}{l|cccccccc}\hline
\textbf{Setting}      & \textbf{Accuracy (\%)} & \textbf{\begin{tabular}[c]{@{}c@{}}Precision (\%)\\ Entailment \end{tabular}} & \textbf{\begin{tabular}[c]{@{}c@{}}Precision (\%)\\ Contradiction\end{tabular}} & \textbf{\begin{tabular}[c]{@{}c@{}}Precision (\%)\\ Neutral\end{tabular}} & \textbf{\begin{tabular}[c]{@{}c@{}}Recall (\%)\\ Entailment\end{tabular}} & \textbf{\begin{tabular}[c]{@{}c@{}}Recall (\%)\\ Contradiction\end{tabular}} & \textbf{\begin{tabular}[c]{@{}c@{}}Recall (\%)\\ Neutral\end{tabular}} \\
\hline
PEGASUS $\rightarrow$ DeBERTAV3 & 33.2 & 64.2& 14.7& 21.5& 37.3& 12.4& 11.9 \\
PEGASUS $\rightarrow$ Llama 2 & 39.5 & 37.4& 23.1& 42.8& 27.6& 24.3& 24.0\\\hline 
T5 large $\rightarrow$ DeBERTAV3 & 34.8 & 62.8& 17.5& 26.2& 33.0& 18.5& 18.2\\
T5 large $\rightarrow$ Llama 2 & 37.2 & 40.2& 32.8& \textbf{48.0}& 30.5& 26.4& 26.8\\\hline  
PRIMERA $\rightarrow$ DeBERTAV3 & 35.9 & 68.2 & 32.7 & 23.7 & 35.8 & 23.8 & 45.1 \\
PRIMERA $\rightarrow$ Llama 2 & 39.2 & 43.5 & 47.2 & 33.1 & 47.2 & 35.5 & 24.9 \\ \hline
MSP(Entail., LM critics) $\rightarrow$ DeBERTAV3 & 36.9 & \textbf{74.3} & 29.3 & 28.3 & 42.5 & 23.9 & \textbf{44.8}\\ 
MSP(Entail. critic) $\rightarrow$ Llama 2 & 42.6 & 41.0 & \textbf{53.7} & 34.6 & 54.8 & 37.8 & 29.6 \\
MSP(Entail., LM critics) $\rightarrow$ Llama 2 & \textbf{46.0} & 49.5 & 49.3 & 34.1 & \textbf{56.4} & \textbf{44.7} & 28.9 \\ \hline
\end{tabular}}
\label{label:multi-news}
\end{table*}

\begin{table*}[th]\tiny
\centering
\caption{Performace of explanation generation. Our system outperforms MOCHEG on equivalent settings. Gold Truthfulness denotes ground truth claim label and System Truthfulness means the predicted claim label.}
\begin{tabular}{l|cccccc}
\hline
\textbf{Setting} & \textbf{ROUGE 1 (\%)} & \textbf{ROUGE 2 (\%)} & \textbf{ROUGE L (\%)} & \textbf{BLEU (\%)} & \textbf{BERTScore (\%)}\\ \hline
MOCHEG w/ Gold Evidence, Gold Truthfulness      & \textbf{45.5}   & \textbf{27.3}   & \textbf{35.4}   & \textbf{21.8}   & \textbf{89.0}   \\
MOCHEG w/ Gold Evidence, System Truthfulness    & 43.8   & 26.3   & 34.1   & 20.8   & 88.8   \\\hline
MOCHEG w/ System Evidence, Gold Truthfulness    & 35.5   & 17.4   & \textbf{26.0}   & \textbf{10.9}   & 87.0   \\
MOCHEG w/ System Evidence, System Truthfulness  & 33.8   & 16.5   & 24.8   & 10.0   & 86.9   \\
MSP(Entail., LM critics) w/ System Evidence, Gold Truthfulness      & \textbf{37.8}   & \textbf{19.4}   & 24.6   & 11.4   & \textbf{88.1}  \\
MSP(Entail., LM critics) w/ System Evidence, System Truthfulness    & 35.1   & 16.3   & 24.9   & 10.6   & 87.5  \\\hline
\end{tabular}
\label{label:explanation_generation}
\end{table*}

We also removed the image component from our model and tested it on the FEVER testing dataset. Using Llama2, our zero-shot entailment prediction accuracy was 62\%. Using PRIMERA's summaries, the most competitive baseline for multidocument summarization, we obtained 42\% on FEVER. Thus, we continue to outperform on this benchmark. Notably, we did not fine-tune the downstream classifier on any benchmark, unlike FEVER's evaluation methods that directly train on labeled data in FEVER. Fine-tuning the final classifier would likely yield better results, but this is orthogonal to our contribution. Our zero-shot evaluation on FEVER, compared to the most competitive multidocument summarization baseline, demonstrates that our approach significantly outperformed (62\% vs. 42\%), underscoring the importance of fact-checking-driven summarization.

\begin{figure}
    \centering
    \includegraphics[width=0.4\textwidth,height=0.4\textwidth,keepaspectratio]{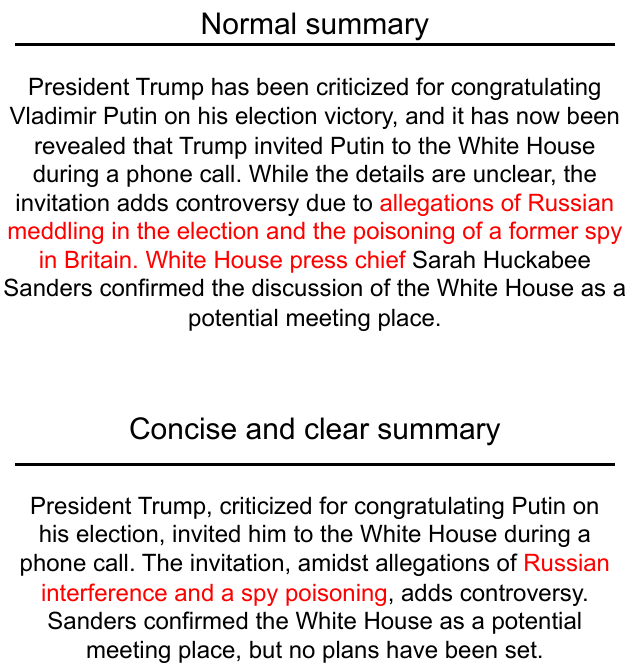}
  \caption{The normal summary is produced by our initial MSP model, while the concise and clear summary is generated using MSP trained with the $r_{quality}$ reward.}
  \label{fig:concise}
\end{figure}

\subsection{Explanation Generation}

\begin{figure*}[t]
  \centering
\includegraphics[width=\textwidth,height=0.6\textwidth,keepaspectratio]{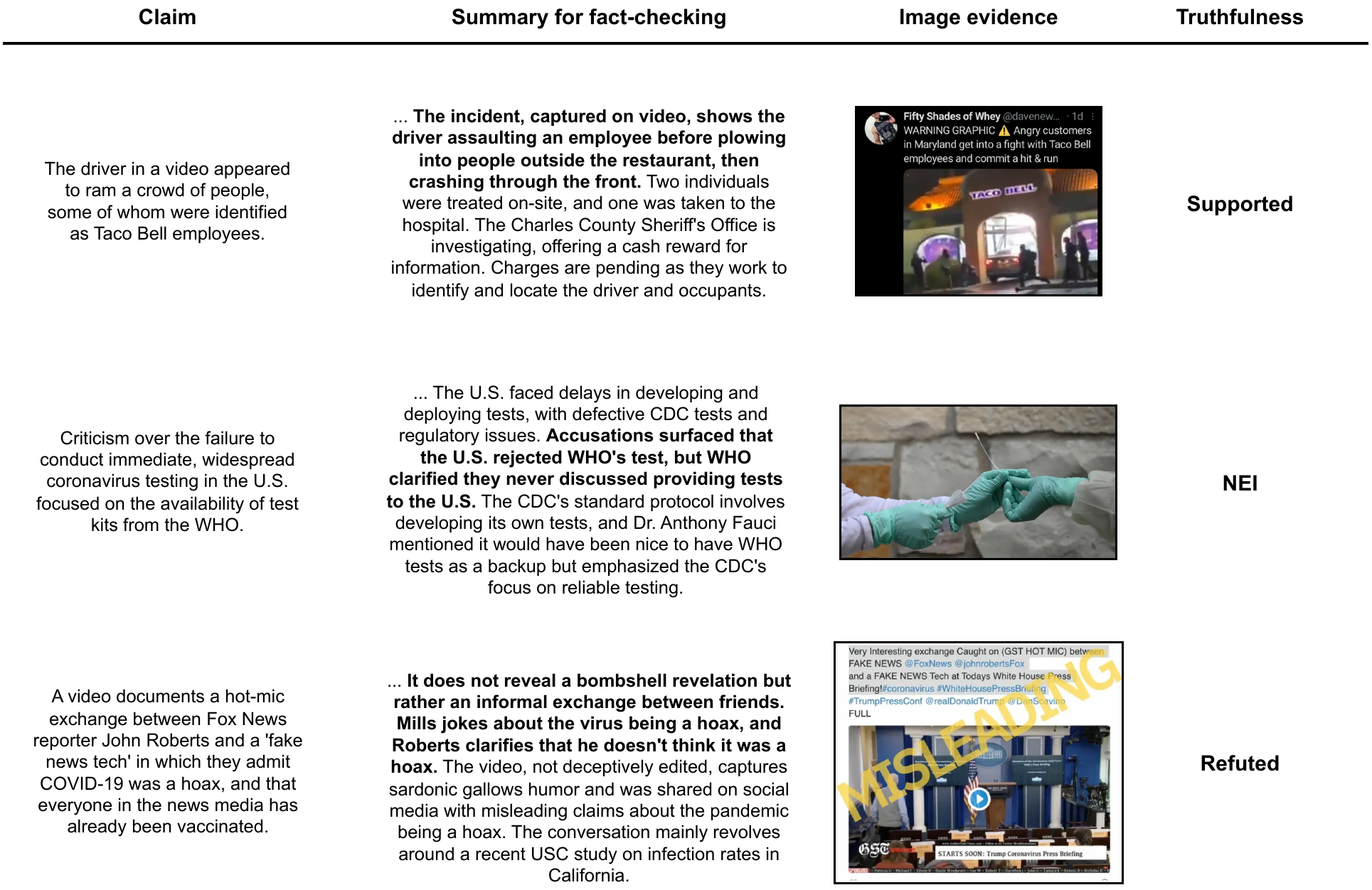}
  \caption{Explanation generation examples of Multimodal Fact-Checking. The Truthfulness column shows gold labels.}
  \label{fig:qualitative}
\end{figure*}

In order to assess the degree to which our generated summaries contain the relevant facts necessary to fact check the generated claims, we measure the ability of a method to generate an~\textit{explanation} of the predicted truthfulness label using our summary. We adopt a methodology similar to~\citet{Yao_2023}, where we consider the input claim $C$, its truthfulness label $Y_C$, and the summary for fact-checking $\{T_1, T_2, ...\}$ generated from MSP.  These components are concatenated into an overall sequence $X$ using a separator </s>. During the training of the rationale generator, we employ the actual truthfulness label of each claim as input. Critically, we do not retrain or fine-tune MSP for this task. In the evaluation phase, we utilize the truthfulness label predicted by the fixed entailment models. 

Following~\citet{Yao_2023}, we utilize BART to generate the ruling statement. Our evaluation metrics include ROUGE~\cite{lin-2004-rouge}, BLEU~\cite{10.3115/1073083.1073135}, and BERTScore~\cite{zhang2020bertscore}. To assess the performance of explanation generation, we compare it with MOCHEG~\cite{Yao_2023}, as shown in Table~\ref{label:explanation_generation}, Figure~\ref{fig:concise} and ~\ref{fig:qualitative}.

We observe that our model outperforms MOCHEG's evidence-retrieval based method (``system evidence'') on the rationale generation task. In our case, ``system evidence'' is our generated summary. We note that MOCHEG's method relies on retrieval from a pool of multimodal documents. The ground truth explanations rely on these sentences and thus may share some phrasing. This gives a slight advantage to MOCHEG's method on some metrics that measure n-gram overlap, whereas our method based on summarization may rephrase the same evidence. Nevertheless, we observe that our system outperforms MOCHEG's generated explanations.

We further observe that our explanations generated using system evidence and system truthfulness outperform MOCHEG's method, which relies on the ground truth truthfulness label on the BERTScore metric. Overall, these results demonstrate that our summarizer, which was not trained for the rationale prediction task, is capturing relevant evidence across modalities in a short summary better than MOCHEG's evidence retrieval-based approach.

%\subsection{Qualitative Results}
%We illustrate our generated summaries for fact-checking in Figure~\ref{fig:qualitative} and ~\ref{fig:concise}. Our findings indicate that our summaries include ample evidence to ascertain the accuracy of the claim label. Moreover, they demonstrate the capability to generate clearer and more concise evidence for the fact-checking task. Whether the truthfulness label is affirmed, marked as NEI, or refuted, we consistently present evidence to assist fact-checkers in determining the veracity of the claim.

%% file: conclusion.tex
We present MetaSumPerceiver (MSP), a summarization model crafted to generate concise and informative summaries specifically tailored for fact-checking claims in intricate multimodal datasets. The model's adaptable architecture can handle varying numbers of documents and input types, encompassing documents, images, and claims, by leveraging a perceiver-based design. We train our model using a RL approach, aiming to produce summaries that are instrumental in verifying the accuracy of claims. Moreover, our reward function is designed to generate more concise and clear summaries, aiding in the verification of diverse claims with the assistance of the LM critic. Our experimental assessments on the MOCHEG and our Multi-News-Fact-Checking datasets highlight MSP's robust performance in claim verification and explanation generation tasks and demonstrate its effectiveness in real-world fact-checking scenarios. This contribution underscores MSP's potential to streamline fact-checking processes in today's multimodal information landscape. Finally, we release the publicly accessible Multi-News-Fact-Checking dataset, aimed at assisting researchers in developing multi-document fact-checking methods.

%% file: limitation.tex
Given the societal importance of fact-checking applications, it is important that the limitations of our method be explored. Our experimental results reveal that the surrogate entailment model often assigns truthfulness labels for entailment even when it struggles to fully grasp the relationship between the claim and the summary with evidence. This issue not only impacts the judgment of the claim label but also affects MSP during training. One potential solution is using a textual entailment model adept at managing this uncertainty or excluding such instances during training. Secondly, Llama 2's claims in the Multi-News-Fact-Checking dataset have certain flaws. Our review suggests that neutral claims might mix consistent and conflicting details. Enhancing our data creation prompts or the prompts used in the second-stage claiming could boost Llama 2's understanding.

Our model, trained on English text and topics from the Multi-News benchmarks, may not perform well in other languages without retraining. Care should be taken to ensure the model is trained on data that closely aligns with the target domain of interest, if possible, to minimize errors. Finally, our model relies on identifying relevant and trusted source documents on which to perform summarization and checking. While this document-level retrieval task is orthogonal to our research, failure to retrieve relevant documents will affect the downstream performance of the fact-checking system. If irrelevant documents are used, even true claims might be wrongly challenged. Thus, approaches should confirm that events and entities in sourced documents are directly related, employing sophisticated methods.

%% file: appendix.tex
\begin{figure*}
    \hspace{-1.7cm}
    \includegraphics[width=1.2\textwidth,height=0.7\textwidth]{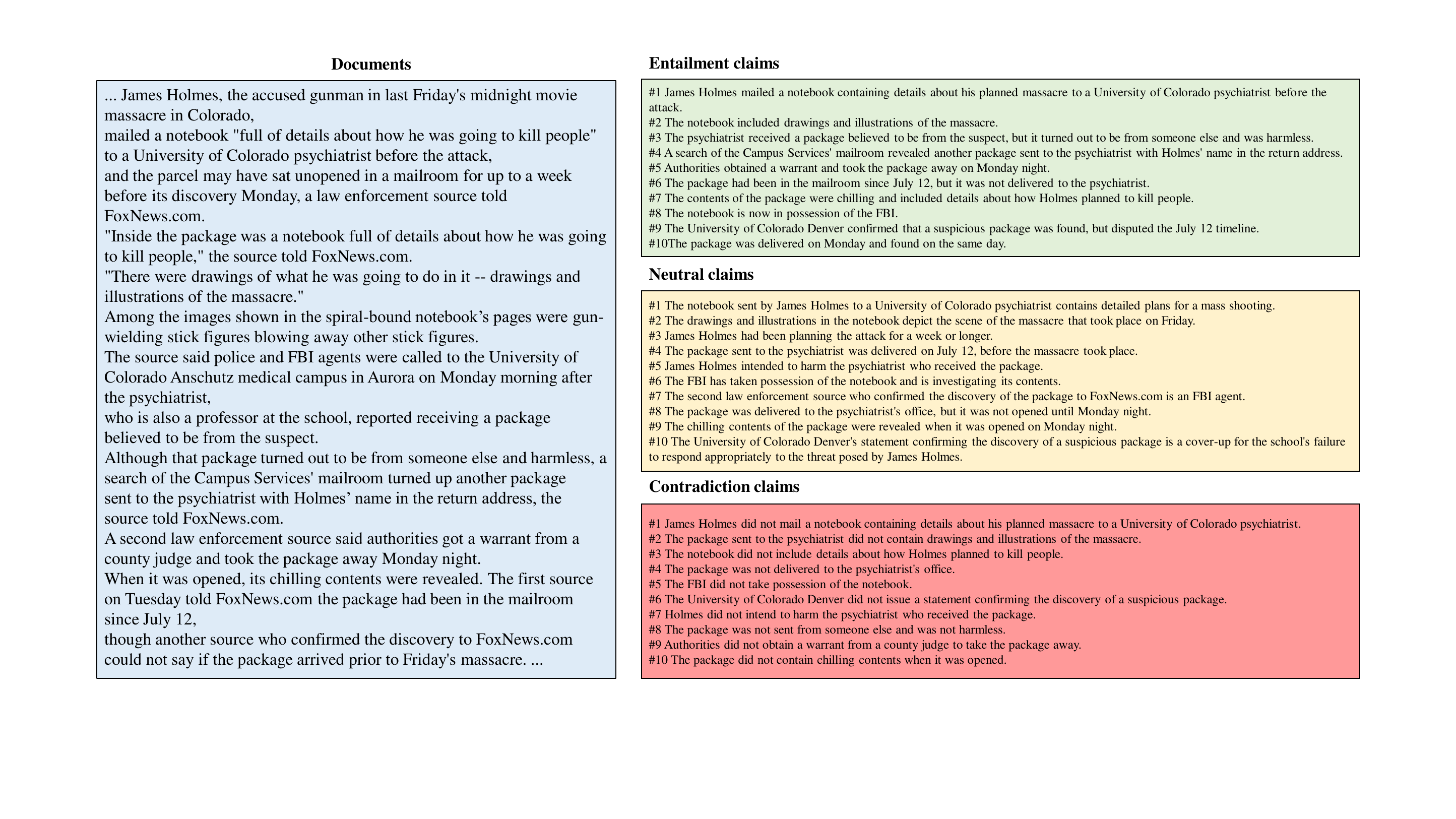}
  \caption{The prompted entailment, neutral, contradiction claims from Llama-2-70b.}
  \label{fig:prompt_ex}
\end{figure*}

\subsection{Prompting For Multi-News-Fact-Checking Dataset}
\label{app:prompts}
In this section, our main goal is to explain the dataset construction process. The following prompts include generating entailment, neutral, and contradiction claims from the Multi-News memorization dataset, ensuring each claim aligns with its corresponding label, and providing clear and concise claims, as depicted in Figure~\ref{fig:prompt_ex}.

\begin{itemize}
    \item \textbf{Prompt for the entailment claims}: Task: You will be provided with a summary of a news article. Your goal is to generate a list of statements derived from the summary. These statements should be definitively true based solely on the information in the summary. Example summary: The unemployment rate dropped to 8.2\% last month, but the economy only added 120,000 jobs, when 203,000 new jobs had been predicted, according to today's jobs report. Reaction on the Wall Street Journal's MarketBeat Blog was swift: "Woah!!! Bad number." The unemployment rate, however, is better news; it had been expected to hold steady at 8.3\%. But the AP notes that the dip is mostly due to more Americans giving up on seeking employment. You will be given a summary of a news article. Your job is to generate a list of entailment claims(true) from the summary. For example, if the summary says job growth was expected to be 100,000 jobs, but only was 80,000 jobs, one simple claim you might write could be "Job growth missed expectations." Please write a numbered list of 10 claims from this summary (numbered 1. through 10.).
    \item \textbf{Prompt for the neutral claims}: Task: You will be provided with a summary of a news article. Your goal is to generate a list of statements derived from the summary. These statements should not be definitively true or false based solely on the information in the summary. In other words, they should be ambiguous and require further investigation or context to determine their accuracy. Example: If the summary mentions that two celebrities are planning to get divorced, you might create a statement suggesting that their divorce might lead to significant financial and legal complications, assuming this information is not explicitly confirmed or denied in the article. Instructions: Review the provided summary. Create 10 statements based on the information in the summary. Each statement should be carefully crafted to be neither definitively true nor false based solely on the summary. Ensure that the truth or falsehood of these statements cannot be logically deduced from the summary alone. Avoid simply rephrasing or restating sentences from the summary; strive for creativity in your statement generation process. Avoid claims using statements like "may" or "could" - your claim should state things as a fact.
    \item \textbf{Prompt for the contradiction claims}: Task: You will be provided with a summary of a news article. Your goal is to generate a list of statements derived from the summary. These statements should be definitively false based solely on the information in the summary. Example: If the summary mentions that a black race car starts up in front of a crowd of people., you might create a statement suggesting that a man is driving down a lonely road assuming this information is explicitly denied in the article. Instructions: Review the provided summary. Create 10 statements based on the information in the summary. Each statement should be carefully crafted to be definitively false based solely on the summary. Avoid simply rephrasing or restating sentences from the summary; strive for creativity in your statement generation process. Avoid claims using statements like "may" or "could" - your claim should state things as a contradiction fact.
    \item \textbf{Prompt for double-check claims}: Task: You will be presented with a set of documents and one claim. Your objective is to discern the claim label based on the information in the documents. The claim labels include entailment, neutral, and contradiction. Entailment signifies that the claim is conclusively true based solely on the documents. The neutral label indicates that the claim should neither be true nor false based on the information provided. The contradiction label implies that the claim is entirely false based on the information presented in the documents.
    \item \textbf{Prompt for the clear and concise claims}: You will be provided a summary that a fact-checker will use for fact-checking a claim. Your task is to evaluate the provided summary using a quality assessment metric that measures whether the summary is factual and written in a clear and concise manner. Good summaries provide facts useful for fact-checking (a general claim) and are short to ease the fact-checkers job. Given a summary, your job is to provide a number from 0 to 1 that indicates your assessment of the quality of the summary. Provide the evaluation score in the format: "The quality score is <score>." The score should range from 0 to 1, where a score of 1 indicates high quality and a score of 0 signifies the lowest quality. The provided summary: <summary>"
\end{itemize}

%\subsection{Refining Multi-News-Fact-Checking Dataset}

%In this section, we present a summary of our dataset evaluation procedures, comprising two primary assessments. We use Llama-2-70b and apply a pre-trained model from the ClaimBuster dataset to determine the check-worthiness of the claims provided to us.

%MSP(Entailment critic, LM critic) $\rightarrow$ DeBERTAV3
%MSP(Entailment critic) $\rightarrow$ Llama 2
%MSP(Entailment critic, LM critic) $\rightarrow$ Llama 2

%In Table~\ref{label:prompt_accuracy_llama2}, we show the performance of Llama-2-70b at predicting the claim label. We demonstrate that Llama-2-70b performs well on entailment claims (78.3\% accuracy) and contradiction claims (74.1\% accuracy). However, its performance is comparatively lower in distinguishing neutral claims, achieving an accuracy of only 64.2\%. This difficulty arises because the neutral category involves identifying specific aspects of a claim that neither entail nor contradict. Similarly, our model faces challenges in distinguishing neutral claims, as evident in the results.

\subsection{Implementation Details}

We used 4 NVIDIA A40 to run our experiments. Our model costs 180 GB and are trained for about 24 runs with a batch size of 256. In the preprocessing and evaluation parts, we use NLTK, ROUGE, and BERTScore. For claim verification, the learning rate $\in\{10^{-4}, 10^{-5}, 10^{-6}\}$ and batch size $\in \{256, 480,512\}$.

\begin{figure*}
\includegraphics[width=1.0\textwidth,height=1.1\textwidth]{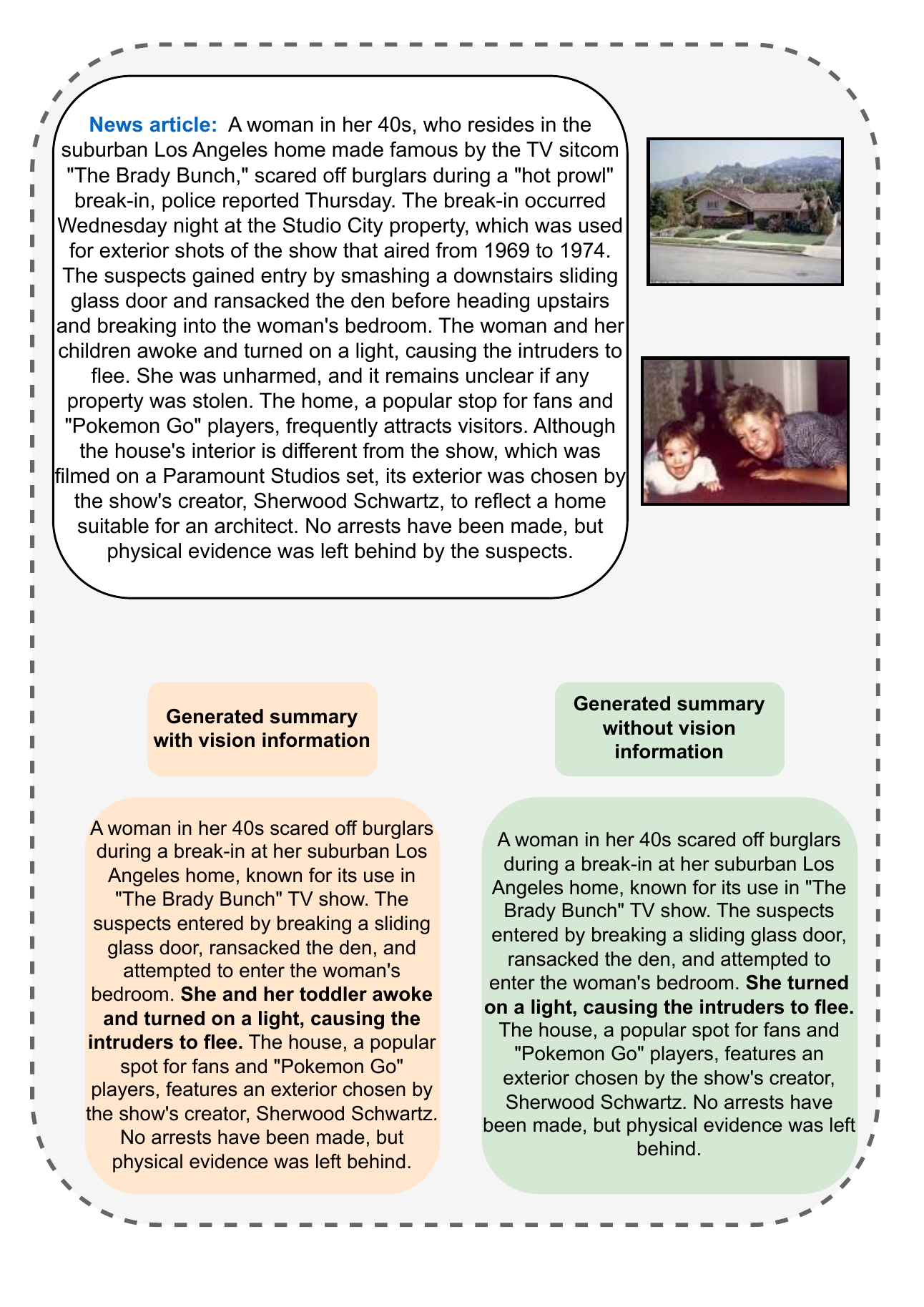}
  \caption{The assessment of generated summaries with and without visual information.}
  \label{fig:p}
\end{figure*}